\pdfoutput=1

\documentclass[11pt]{article}

\usepackage[final]{acl}
\usepackage{times}
\usepackage{latexsym}

\usepackage[T1]{fontenc}

\usepackage[utf8]{inputenc}

\usepackage{microtype}

\usepackage{inconsolata}

\usepackage{graphicx}
\usepackage{tabularx, colortbl}
\usepackage{soul}
\usepackage{booktabs}
\usepackage{makecell}
\usepackage{amssymb}
\usepackage{mathtools}
\usepackage{multirow,longtable,array}

\newcommand{\orte}{\textbf{\textit{ORTE}}}
\newcommand{\orse}{\textbf{\textit{ORSE}}}
\newcommand{\orc}{\textbf{\textit{ORC}}}
\newcommand{\openie}{OpenIE}

\definecolor{lightred}{HTML}{E8CDDA}
\definecolor{lightblue}{HTML}{B0C6FF}
\definecolor{lightyellow}{HTML}{FFD67B}

\title{A Survey on Open Information Extraction from Rule-based Model to Large Language Model}

\author{
 \textbf{Pai Liu\textsuperscript{1,2,\thanks{These authors contributed equally to this work.}}},
 \textbf{Wenyang Gao\textsuperscript{1,\footnotemark[1]}},
 \textbf{Wenjie Dong\textsuperscript{3,\footnotemark[1]}},
 \textbf{Lin Ai\textsuperscript{4,\footnotemark[1]}},
\\
 \textbf{Ziwei Gong\textsuperscript{4,\footnotemark[1]}},
 \textbf{Songfang Huang\textsuperscript{5}},
 \textbf{Zongsheng Li\textsuperscript{6}},
 \textbf{Ehsan Hoque \textsuperscript{2}}
\\
 \textbf{Julia Hirschberg\textsuperscript{4}},
 \textbf{Yue Zhang\textsuperscript{1,\thanks{The corresponding author.}}},
\\
\\
 \textsuperscript{1}Westlake University,
 \textsuperscript{2}University of Rochester,
 \textsuperscript{3}Zhejiang University
\\
 \textsuperscript{4}Columbia University,
 \textsuperscript{5}Alibaba DAMO Academy,
 \textsuperscript{6}Northeastern University
\\
\textsc{\href{mailto:email@domain}{zhangyue@westlake.edu.cn}}
}

\begin{document}
\maketitle

\begin{abstract}

Open Information Extraction ({\openie}) represents a crucial NLP task aimed at deriving structured information from unstructured text, unrestricted by relation type or domain. This survey paper provides an overview of {\openie} technologies spanning from 2007 to 2024, emphasizing a chronological perspective absent in prior surveys. It examines the evolution of task settings in {\openie} to align with the advances in recent technologies. The paper categorizes {\openie} approaches into rule-based, neural, and pre-trained large language models, discussing each within a chronological framework. Additionally, it highlights prevalent datasets and evaluation metrics currently in use. 
Building on this extensive review, this paper systematically reviews the evolution of task settings, data, evaluation metrics, and methodologies in the era of large language models, highlighting their mutual influence, comparing their capabilities, and examining their implications for open challenges and future research directions.

\end{abstract}
\section{Introduction}
\label{sec:intro}

\begin{figure}[tbhp!]
    \centering
    \includegraphics[scale=0.1]{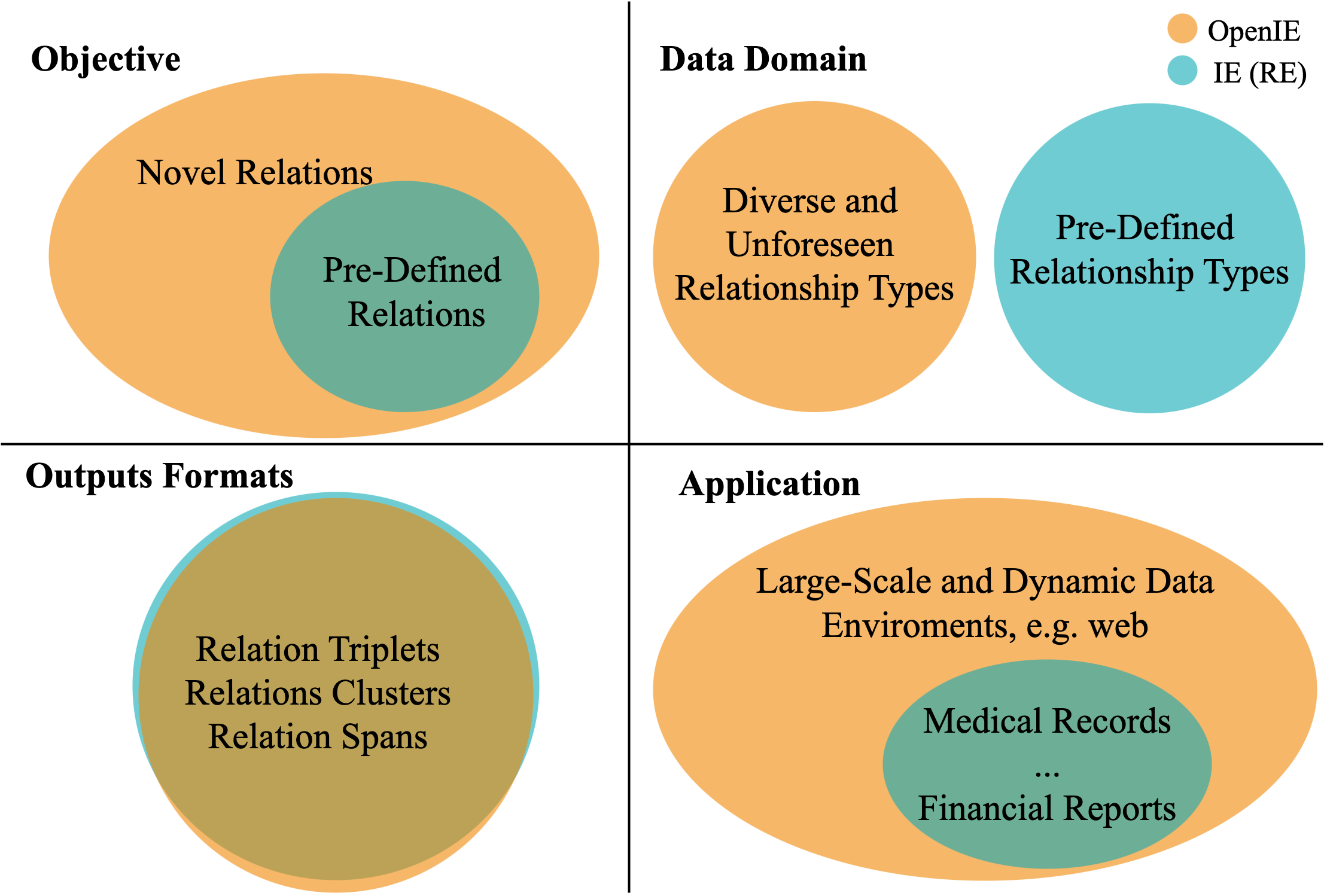}
    \caption{Comparison of {\openie} and standard relation extraction.}
    \label{fig:oie_vs_re}
    \vspace*{-4mm}
\end{figure}

Open Information Extraction ({\openie}) aims to extract structured information from unstructured text sources \cite{Niklaus2018ASO}, typically outputting relationships as triplets $(arg_1, rel, arg_2)$. As illustrated in Figure \ref{fig:oie_vs_re}, unlike standard IE, which relies on predefined categories to identify relationships, {\openie} operates without such constraints, enabling the extraction of diverse and unforeseen relations. This flexibility makes {\openie} especially valuable for rapidly evolving Natural Language Processing (NLP) tasks such as question answering, search engines, and knowledge graph completion \cite{han2020more}, as well as for handling large-scale and dynamic data sources like web data.

Since its inception in 2007, the field of {\openie} has witnessed continual advancements. Initially utilizing basic linguistic tools, {\openie} models have progressively integrated more complex syntactic and semantic features, while preserving the intuitive task of directly extracting relational triplets from text. The advent of neural models in 2019 marks a paradigm shift for {\openie} research, where systems employing Transformer-based architectures like BERT \cite{Devlin2019BERTPO} significantly enhance feature extraction capabilities. To accommodate the technological shift, a variety of methods and task settings have evolved within diversified {\openie} approaches.

The emergence of Large Language Models (LLMs) in 2023 has marked another revolutionary phase, steering {\openie} toward generative information extraction. 
The robust generalization abilities of these models not only advance the technical prowess of {\openie} systems but also facilitate the convergence of methods and task settings -- revisiting the original, straightforward \textbf{\textit{text $\rightarrow$ relational triplet}} format. This transition fosters potential integration with standard IE tasks, pointing toward a promising future where extraction tasks are tackled through a unified, multi-task approach. 


As a result, there has been a decline in {\openie} research in the LLM era. \textit{Is {\openie} research going to its end? How can traditional {\openie} research inspire IE research in the LLM era?} Previous surveys largely focus on pre-LLM era models or limit their scope to methodological insights \cite{overviewOIE2014, vo2018open, ZOUAQ2017228, Glauber2018ASM, Niklaus2018ASO, zhou2022survey}. While recent studies \cite{xu2023large} delve into information extraction in the LLM era, they largely bypass {\openie}, concentrating instead on standard IE tasks. We aim to bridge this gap by providing a holistic survey of the {\openie} field from a chronological view, addressing the two research questions above.

From a chronological perspective, we summarize all task settings (Section \ref{sec:tasksettings}), data (Section \ref{sec:data}), evaluation metrics (Section \ref{sec:eva}), and mainstream methods (Section \ref{sec:methodologies}) before and after LLM era. We use a single table to summarize mainstream methods and results from different periods. We emphasize the co-evolution between models and task settings, and the various sources of information used to address Open challenges. Based on this, we compare the ideas and relative strengths and weaknesses of large models and traditional models (Section \ref{sec:discuss_coevolution}), review the impact of large language models on open information extraction and traditional methods (Section \ref{subsec:discuss_llm}), and explore future directions (Section \ref{sec:futuredir}).

\section{Task Settings} 
\label{sec:tasksettings}

\vspace{-1pt}
\begin{figure*}[t]
    \centering
    \includegraphics[width=0.92\textwidth]{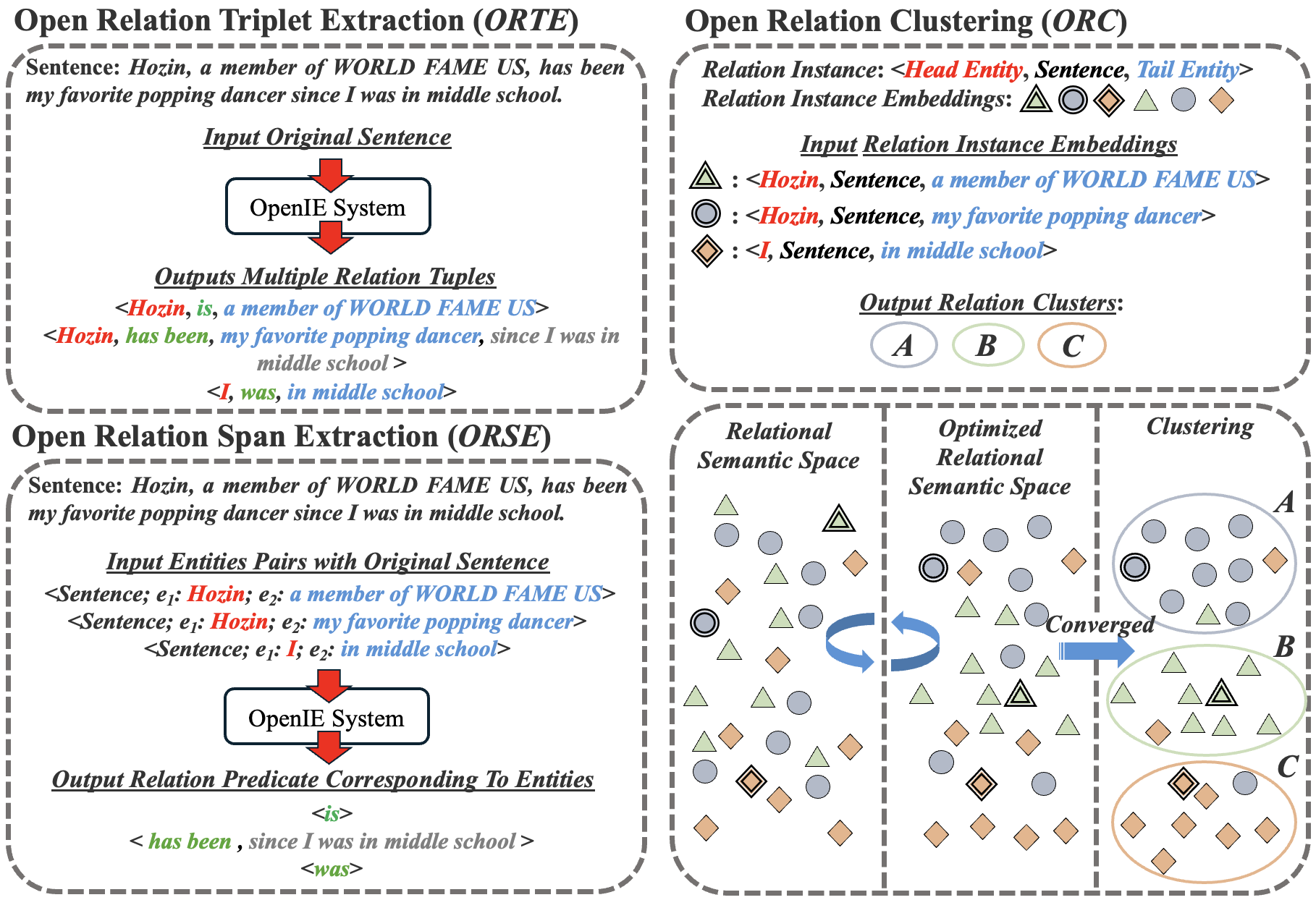}
    \caption{An overview of workflow processes in {\openie} task settings. ORTE aims to extract all n-ary relation tuples in the input, . ORSE finds relational spans according to previously extracted subjects and Objects. ORC pairs the input sentence with different subjects and objects within the sentence to form relation instances, relation instances are iteratively optimized in a supervised, unsupervised, or semi-supervised manner, and after the representations converge, clustering is performed. Objects of the same color indicate that they belong to the same relation cluster. To facilitate observation, we have bolded the borders of the three examples in the figure. Each circle represents the relation instances clustered into the same class by the clustering algorithm. }
    \label{fig:taskSettingAll}
    \vspace*{-3mm}
\end{figure*}
\vspace{-2pt}

We categorize {\openie} task settings into three groups: Open Relation Triplet Extraction ({\orte}),  Open Relation Span Extraction ({\orse}) and Open relation clustering ({\orc}). {\orte} is the classic task setting, while {\orse} and {\orc} settings are variations developed to cater to diverse models with the advancement of NLP techniques. For all three task settings, openness is shown in the absence of restraints on relation types. 
Figure \ref{fig:taskSettingAll} depicts the workflow for each task setting.

\noindent 
\textbf{\textit{ORTE Task: Text $\rightarrow$ Relational Triplet}} \\
\citet{banko2007open} initially defines open information extraction as an unsupervised task that automatically extracts $(entity_1, relation, entity_2)$ triplets from a vast corpus of unstructured web text, where $entity_1$, $entity_2$ and $relation$ consist of selected words from input sentences. Although the term \textit{triplet} is more commonly used, the actual extraction tasks are not always limited to triplets and can involve more diverse n-ary relations, such as condition, temporal information, etc. This task setting, irrespective of the learning method or the forms of input and output, represents the most idealized configuration. 


\noindent\textbf{\textit{ORSE Task: Entities + Text $\rightarrow$ Relation Span}}\\
Different from the first setting, open relation span extraction finds relational spans according to previously extracted predicates and entities, aiming to partition complex tasks into easier ones to improve model performance. 
However, it should be clear that errors in entity extraction steps can accumulate in two-stage pipelines. See Open Relation Extraction ({\orse}) in Fig.\ref{fig:taskSettingAll} for an example.

\noindent\textbf{\textit{ORC Task: Entities + Text$\rightarrow$ Clustering without Explicit Relation Span or Label}}\\
Open relation clustering ({\orc}), also known as open relation extraction, clusters relation instances $(h,t,s)$, where $h$ and $t$ denote head entity and tail entity respectively, and $s$ denotes the sentence corresponding to two entities. Different from the {\orte}, {\orc} does not extract relation from text but uses text between two entities to represent the relation. Clustering similar relations is a step forward in labeling specific relations to each relation instance. 
These task settings outlined above are distinctly characterized by era-specific traits and methodologies, further discussed in Section \ref{sec:methodologies}. 

Summarized in Table \ref{tab:evaluation}, ORTE efficiently identifies all relations in one step, reducing error and ensuring consistency, but struggles with complex implementation, overlapping relations, and large data requirements. ORSE simplifies individual models, handling nested relations by focusing on specific subtasks, but is prone to error propagation and is computationally costly. ORC clusters diverse relation expressions to aid generalization, but faces scalability challenges due to intensive similarity computations and potential loss of specificity.




\section{Datasets} 
\label{sec:data}

Table~\ref{tab:dataset} lists some popular and promising {\openie} datasets grouped by their creating methods. 

\textbf{Question Answering (QA) derived datasets} are converted from other crowd-sourced QA datasets. OIE2016 \cite{Stanovsky2016CreatingAL}, one of the most popular {\openie} benchmarks, leverages QA-SRL \cite{He2015QuestionAnswerDS} annotations. Additional datasets extend from OIE2016, such as AW-OIE \cite{Stanovsky2018SupervisedOI}, Re-OIE2016 \cite{Zhan2020SpanMF} and CaRB \cite{Bhardwaj2019CaRBAC}. LSOIE \cite{Solawetz2021LSOIEAL}, is created by converting the QA-SRL 2.0 dataset \cite{FitzGerald2018LargeScaleQP} to a large-scale {\openie} dataset,  which claims to be 20 times larger than the next largest human-annotated {\openie} dataset.


\textbf{Crowdsourced datasets} are created from direct human annotation, including WiRe57 \cite{Lchelle2019WiRe57A}, SAOKE dataset \cite{Sun2018LogicianAU}, and BenchIE dataset \cite{Gashteovski2021BenchIEOI}. 

\textbf{Knowledge Base (KB) derived datasets} are established by aligning triplets in KBs with text in the corpus. 
Several works \cite{Mintz2009DistantSF,Yao2011StructuredRD} have aligned the New York Times corpus \cite{sandhaus2008new} with Freebase \cite{Bollacker2008FreebaseAC} triplets, resulting in several variations of the same dataset, NYT-FB. Others are created by aligning relations of given entity pairs \cite{ElSahar2018TRExAL}, such as TACRED \cite{Zhang2017PositionawareAA}, FewRel \cite{Han2018FewRelAL}, T-REx \cite{ElSahar2018TRExAL},  T-REx SPO and T-REx DS \cite{hu2020selfore}. COER \cite{Jia2018ChineseOR}, a large-scale Chinese KB dataset, is automatically created by an unsupervised open extractor.

\textbf{Instruction-based datasets} transform IE tasks into  instruction-following tasks to harness the capabilities of LLMs. Strategies include integrating existing IE datasets into a unified-format \cite{wang2023instructuie, lu-etal-2022-unified}, and deriving others from Wikidata and Wikipedia such as \textsc{InstructOpenWiki} \cite{lu2023pivoine}, \textsc{InstructIE} \cite{gui2023instructie}, and Wikidata-OIE \cite{wang-etal-2022-ielm}. 

Overall, KB derived datasets are mostly used in {\orc} task settings, whereas QA derived, crowd-sourced, and instruction-based datasets are usually used in {\orte} and {\orse} task settings. We provide more detailed descriptions in Appendix \ref{appendix:datasets}.

\vspace{-2pt}
\begin{table}[t]
\centering
\small
\resizebox{0.47\textwidth}{!}{

\begin{tabular}{@{}lrrr@{}}
\toprule
\textbf{Dataset}              & \textbf{\#Tuple}           & \textbf{Domain}   & \textbf{Task}  \\ \midrule
\multicolumn{3}{c}{\textbf{\textit{QA Derived}}} \\
OIE2016 \citeyearpar{Stanovsky2016CreatingAL}   & 10,359            & Wiki, Newswire  & ORTE/ORSE                   \\
Re-OIE2016 \citeyearpar{Zhan2020SpanMF}  & NR                  & Wiki, Newswire         & ORTE/ORSE            \\
CaRB \citeyearpar{Bhardwaj2019CaRBAC}      & NR                  & Wiki, Newswire    & ORTE/ORSE                 \\
AW-OIE \citeyearpar{Stanovsky2018SupervisedOI}  & 17,165            & Wiki, Wikinews      & ORTE/ORSE               \\ 
LSOIE-wiki \citeyearpar{Solawetz2021LSOIEAL} & 56,662            & Wiki, Wikinews        & ORTE/ORSE             \\
LSOIE-sci \citeyearpar{Solawetz2021LSOIEAL} & 97,550            & Science          & ORTE/ORSE                  \\ \midrule
\multicolumn{3}{c}{\textbf{\textit{Crowdsourced}}} \\
WiRe57 \citeyearpar{Lchelle2019WiRe57A} & 343               & Wiki, Newswire       & ORTE/ORSE              \\
$\textrm{SAOKE}^{{zh}}$ \citeyearpar{Sun2018LogicianAU} & NR                   & Baidu Baike        & ORC                \\
$\textrm{BenchIE}^{{en}}$ \citeyearpar{Gashteovski2021BenchIEOI} & 136,357           & Wiki, Newswire        & ORTE/ORSE             \\
$\textrm{BenchIE}^{{de}}$ \citeyearpar{Gashteovski2021BenchIEOI} & 82,260            & Wiki, Newswire       & ORTE/ORSE              \\
$\textrm{BenchIE}^{{zh}}$ \citeyearpar{Gashteovski2021BenchIEOI} & 5,318             & Wiki, Newswire       & ORTE/ORSE              \\ \midrule
\multicolumn{3}{c}{\textbf{\textit{KB Derived}}} \\
NYT-FB \citeyearpar{sandhaus2008new,Bollacker2008FreebaseAC,Mintz2009DistantSF,Yao2011StructuredRD} & 39,000               & NYT, Freebase           & ORC          \\
TACRED \citeyearpar{Zhang2017PositionawareAA} & 119,474                  & TAC KBP              & ORC                \\
FewRel \citeyearpar{Han2018FewRelAL} & 70,000               & Wiki, Wikidata         & ORC            \\
T-REx \citeyearpar{ElSahar2018TRExAL} & 11M                  & Wiki, Wikidata           & ORC                    \\
$\textrm{COER}^{{zh}}$ \citeyearpar{Jia2018ChineseOR} & 1M & \makecell{Baidu Baike,\\Chinese news} & ORC\\ \midrule
\multicolumn{3}{c}{\textbf{\textit{Instruction-Based}}} \\
\textsc{InstructOpenWiki} \citeyearpar{lu2023pivoine}  & 19M & Wiki, Wikidata & ORTE/ORSE\\
Wikidata-OIE \citeyearpar{wang-etal-2022-ielm}  & 27M & Wiki, Wikidata & ORTE/ORSE\\
\bottomrule
\end{tabular}}

\caption{Statistics of popular {\openie} datasets. "NR" stands for "Not Reported". Non-English datasets are indicated with superscripts. The Task column indicates the types of tasks the data can be used for.}
\label{tab:dataset}
\end{table}
\vspace{-2pt}

\section{Evaluation} 
\label{sec:eva}

\vspace{-2pt}
\begin{table*}[t!]
\centering
\small
\begin{tabular}{c|c|c|p{8cm}} 
\toprule
\textbf{Task Setting} & \textbf{Datasets} & \textbf{Evaluation Metrics} & \textbf{Advantages / Disadvantages} \\ 
\midrule
\multirow{2}{*}{\makecell[c]{ORTE}} & \makecell[c]{OIE2016\\ Re-OIE2016\\ CaRB} & \makecell[c]{Precision\\ Recall\\ F1\\ AUC} & \makecell[l]{\textbf{Advantages:} Efficiency; Reduced Error Propagation\\  Consistency\\ \textbf{Disadvantages:} Difficulty in Handling Overlapping Relations;\\ Handling complex relations } \\
\midrule
\multirow{2}{*}{\makecell[c]{ORSE}} & \makecell[c]{OIE2016\\ Re-OIE2016\\ CaRB} & \makecell[c]{Precision\\ Recall\\ F1\\ AUC} & \makecell[l]{\textbf{Advantages:} Control Over Intermediate Steps; Handling of \\Overlapping Relations; Modular approach allows for specialized \\improvements; More Explainable\\ \textbf{Disadvantages:} Error propagation; Increased Computational \\Cost; Complexity in Integration} \\ 
\midrule
\multirow{2}{*}{\makecell[c]{ORC}} & \makecell[c]{NYT-FB\\ FewRel\\ TACRED\\ T-REx} & \makecell[c]{B\textsuperscript{3}\\ V-measure\\ ARI} & \makecell[l]{\textbf{Advantages:} Flexibility; Generalization; Lower dependency on \\annotated data; Knowledge Organization\\ \textbf{Disadvantages:} Lack of explicit relation labels, Ambiguity in \\clusters; Scalability Issues; Loss of Specificity} \\
\bottomrule
\end{tabular}

\caption{Overview of task settings in Open Information Extraction (OpenIE), detailing frequently used datasets, typical evaluation metrics, and a summary of key advantages and disadvantages associated with each task.}
\label{tab:evaluation}
\end{table*}
\vspace{-2pt}

Evaluation metrics for {\openie} models vary by task setting. In the {\orte} and {\orse} settings, models are assessed using precision, recall, F1 score, and AUC, potentially employing various scoring functions. In the {\orc} setting, performance is evaluated using $B^3$ \cite{Bagga1998EntityBasedCC}, V-measure \cite{Rosenberg2007VMeasureAC}, and ARI \cite{Hubert1985ComparingP}.

To compare the extracted and golden triplets, various datasets employ different matching strategies, typically categorized into \textbf{token-level} and \textbf{fact-level} scorers. Token-level scorers focus on individual tokens to ensure precision and semantic accuracy, accommodating linguistic variability \cite{Stanovsky2016CreatingAL}, enhancing conciseness \cite{Lchelle2019WiRe57A}, and adapting to complex model outputs like those from LLMs \cite{han2023information}. Fact-level scorers assess the informational faithfulness of extractions to ensure reliable knowledge extraction, validating semantic and information integrity \cite{Sun2018LogicianAU, Gashteovski2021BenchIEOI, li2023evaluating} to enhance {\openie} evaluations comprehensively. Further details are discussed in Appendix \ref{appendix:evaluation}. 

From the perspective of task formulation, token-level scorers are better suited for open relation span extraction ({\orse}), where outputs are succinct, and labeling models in open relation triplet extraction ({\orte}), whose outputs are precise tokens derived from the inputs. Conversely, fact-level scorers are more appropriate for generative models in {\orte}, particularly LLMs, whose outputs exhibit diversity and necessitate semantic evaluation.

\begin{table*}[t]
\centering
\small

\resizebox{1.0\textwidth}{!}{
\begin{tabular}{l|c|l|cc|cc|cc|ccc|ccc}

\toprule
 & & & \multicolumn{2}{c|}{\textbf{ OIE16 }} & \multicolumn{2}{c|}{\textbf{ Re-OIE16 }} & \multicolumn{2}{c|}{\textbf{ CaRB }}  & \multicolumn{3}{c|}{\textbf{ FewRel }} & \multicolumn{3}{c}{\textbf{ TACRED }} \\
& \textbf{Category}& \textbf{Representative Approach} & F1 & AUC & F1 & AUC & F1 & AUC & ARI & \textit{$B^{3}$} & V& ARI & \textit{$B^{3}$} & V\\
\midrule
\rowcolor{lightred}
&-& \textsc{Ollie} \cite{schmitz2012open}& 38.6 & 20.2 & 49.5 & 31.3 & 41.1 & 22.4 & -& -& -& -& -& -\\
 \rowcolor{lightred}
 &- & ClausIE \cite{del2013clausie}& 58.0 & 36.4 & 64.2 & 46.4 & 44.9 & 22.4 & -& -& -& -& -& -\\ 
 \rowcolor{lightred}
 &- & \textsc{OpenIE4} \cite{mausam2016open} & 58.8 & 40.8 & 68.3 & 50.9 & 51.6 & 29.5 & -& -& -& -& -& -\\
\rowcolor{lightred}
\multirow{-4}{*}{\makecell[c]{\textbf{Pre-Neural ({\orte})}\\2007 - 2018}} &- & PropS \cite{Stanovsky2016CreatingAL}& 54.4 & 32.0 & 64.2 & 43.3 & 31.9 & 12.6 & -& -& -& -& -& -\\

\midrule\midrule

\rowcolor{lightred}
& SL & RnnOIE \cite{Stanovsky2018SupervisedOI} & 62.0 & 48.0 & -& -& 49.0 & 26.1 & -& -& -& -& -& -\\
 \rowcolor{lightred}
 & SL & OpenIE6 \cite{Kolluru2020OpenIE6IG}& -& -& -& -& 52.7 & 33.7 & -& -& -& -& -& -\\
 \rowcolor{lightred}
 & Span & SpanOIE \cite{Zhan2020SpanMF}& 69.4 & 49.1 & 77.0 & 65.8 & 48.5 & -& -& -& -& -& -& -\\
 \rowcolor{lightred}
 & S2S & IMoJIE \cite{Kolluru2020IMoJIEIM}& -& -& -& -& 53.5 & 33.3 & -& -& -& -& -& -\\
 \rowcolor{lightred}
 & SL & MacroIE \cite{Bowen2021MaximalCB}& -& -& -& -& 54.8 & 36.3 & -& -& -& -& -& -\\
 \rowcolor{lightred}
 & SL & DetIE$_{LSOIE}$ \cite{Vasilkovsky2022detie}& -& -& -& -& 43.0 & 27.2 & -& -& -& -& -& -\\
 \rowcolor{lightred}
 & SL & DetIE$_{IMoJIE}$ \cite{Vasilkovsky2022detie}& -& -& -& -& 52.1 & 36.7 & -& -& -& -& -& -\\
 \rowcolor{lightred}
 \multirow{-8}{*}{\makecell[c]{\textbf{Neural Era ({\orte})}\\2018 - 2022}} & SL & SMiLe-OIE \cite{DBLP:conf/emnlp/DongSK022}& -& -& -& -& 53.8 & 34.9 & -& -& -& -& -& -\\
 
\hline
\rowcolor{lightblue}
& SL & $\textrm{Multi}^{2}\textrm{OIE}$ \cite{Ro2020Multi2OIEMO}& -& -& 83.9 & 74.6 & 52.3 & 32.6 & -& -& -& -& -& -\\
\rowcolor{lightblue}
 & S2S & \textsc{Gen2OIE} \cite{Kolluru2022AlignmentAugmentedCT} & -& -& -& -& 54.4& 32.3& -& -& -& -& -& -\\
 \rowcolor{lightblue}
 & S2S &   \textsc{Gen2OIE} (label-rescore)& -& -& -& -& 54.5& \textbf{38.9}& -& -& -& -& -& -\\
 \rowcolor{lightblue}
 & Graph & OIE@OIA \cite{Wang2022OIEOIAAA}& \textbf{71.6} & \textbf{54.3} & \textbf{85.3} & \textbf{76.9} & 51.1 & 33.9 & -& -& -& -& -& -\\
 \rowcolor{lightblue}
 & Graph & DragonIE \cite{yu2022generalized}& -& -& -& -& 55.1& 36.4& -& -& -& -& -& -\\
  \rowcolor{lightblue}
 & Graph & ChunkOIE(SaC-OIA-SP) \cite{Dong2023OpenIE}& -& -& -& -& 53.6& 35.5& -& -& -& -& -& -\\ 
 \rowcolor{lightblue}
 \multirow{-7}{*}{\makecell[c]{\textbf{Neural Era ({\orse})}\\2018 - 2022}} & Graph & ChunkOIE(SaC-CoNLL) & -& -& -& -& 53.2& 34.7& -& -& -& -& -& -\\ 
 
\hline
\rowcolor{lightyellow}
& Semi & RSN \cite{wu2019open} & -& -& -& -& -& -& 45.3& 58.9& 70.8& 45.9& 63.1& 64.3\\ 
\rowcolor{lightyellow}
 & Semi & RSN-CV \cite{wu2019open} & -& -& -& -& -& -& 54.2& 63.8& 72.4& -& -& -\\
\rowcolor{lightyellow}
 & Un & SelfORE \cite{hu2020selfore}& -& -& -& -& -& -& 64.7 & 67.8 & 78.3 & 44.7 & 54.1 & 61.9 \\
\rowcolor{lightyellow}
 & Semi & RSN-BERT \cite{Zhao2021ARC} & -& -& -& -& -& -& 53.2& 70.9& 78.1& 75.6& 83.4& 85.9\\
\rowcolor{lightyellow}
 & Semi & RoCORE \cite{Zhao2021ARC} & -& -& -& -& -& -& 70.9& 79.6& 86& 81.2& 86& 88.8\\ 
\rowcolor{lightyellow}
   & Semi & OHRE \cite{zhang2021open} & -& -& -& -& -& -& 64.2& 70.5& 76.7& -& -& -\\
\rowcolor{lightyellow}
 & Semi & MatchPrompt \cite{wang-etal-2022-matchprompt}& -& -& -& -& -& -& 66.5 & 72.3 & 82.2 & 75.3 & 83.0 & 84.5 \\
\rowcolor{lightyellow}
 & Un & PromptORE \cite{genest2022promptore} & -& -& -& -& -& -& 43.4& 48.8& 71.8& -& -& -\\ 
\rowcolor{lightyellow}
 & Semi & CaPL \cite{duan-etal-2022-cluster} & -& -& -& -& -& -& \textbf{79.4}& \textbf{81.9}& \textbf{88.9}& \textbf{82.9}& \textbf{87.3}& \textbf{89.8}\\ 
\rowcolor{lightyellow}
\multirow{-10}{*}{\makecell[c]{\textbf{Neural Era ({\orc})}\\2018 - 2022}} & Semi & ASCORE \cite{zhao-etal-2023-actively} & -& -& -& -& -& -& 67.6& 73.5& 83.5& 78.1& 78& 83.1\\ 
 
 \midrule\midrule

\rowcolor{lightred}
& 0-shot & IELM GPT-2$_{XL}$ \cite{wang-etal-2022-ielm} & - & - & 35.0 & - & 22.7 & -  & - & - & - & - & - & - \\
\rowcolor{lightred}
 & few-shot & GPT-3.5-\textsc{Turbo} ICL \cite{ling2023improving} & 65.1 & - & 67.9 & - & 52.1 & - & - & - & - & - & - & - \\ 
\rowcolor{lightred}
\multirow{-3}{*}{\makecell[c]{\textbf{Transition Era to}\\ \textbf{LLM ({\orte})}\\2022 -}} & few-shot & ChatGPT $n$-shot \cite{qi2023mastering} & - & - & - & - & \textbf{55.3} & - & - & - & - & - & - & - \\
\bottomrule
\end{tabular}
}

\caption{Performance of {\openie} models. For \textit{$B^{3}$} and V measures, F1 scores are reported. Rows filled with colors represent models of different task settings: \colorbox{lightred}{\rule{0pt}{6pt}\rule{6pt}{0pt}} = {\orse}, \colorbox{lightblue}{\rule{0pt}{6pt}\rule{6pt}{0pt}} = {\orse}, \colorbox{lightyellow}{\rule{0pt}{6pt}\rule{6pt}{0pt}} = {\orc}. \textit{SL} stands for sequence labeling, \textit{S2S} denotes sequence to sequence, \textit{Span} represents a span-based method, \textit{Graph} signifies a graph-based method, \textit{Semi} refers to semi-supervised learning, and \textit{Un} refers to unsupervised learning.
}
\label{tab:oiePerformancesFull}
\end{table*}

\section{A Chronological Review of Mainstream Methods}
\label{sec:methodologies}
The research approaches for Open IE have undergone three significant changes along with technological advancements. We categorize these periods into three eras: the pre-neural era, dominated by rule-based and statistic-based methods; the neural model era, primarily based on neural networks; and the LLMs era, characterized by the use of LLMs. Chronologically, we will discuss the key models and methods from each period and explore their connections, as depicted in Figure \ref{fig:method_chron_view}. 
More details on model implementation provided in Appendix.\ref{sec:methodAPP}  
\begin{figure}[t]
    \centering
    \includegraphics[width=\columnwidth]{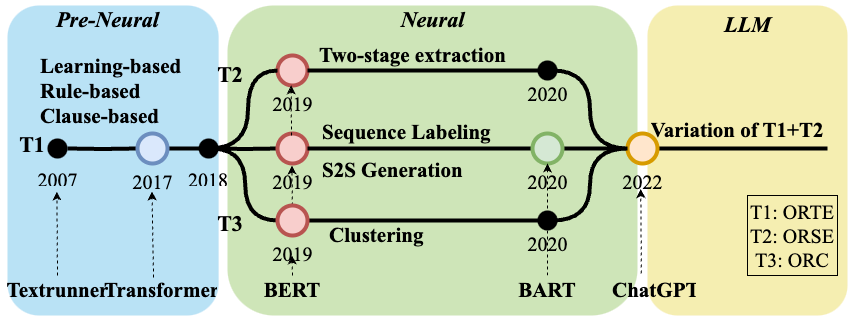}
    \caption{Chronological overview of Open IE methods.}
    \label{fig:method_chron_view}
\end{figure}
\subsection{Pre-neural Model Era}
\label{sec:rule}

In the beginning, {\openie} systems were developed to create a universal model capable of extracting relation triplets through shallow features, such as Part-of-Speech (POS) that do not have lexical information, for instance, characterizing a verb based on its context. Traditional machine learning models, such as Naive Bayes \cite{rish2001empirical} and Conditional Random Field \cite{sutton2012introduction}, are used to train on shallow features \cite{yates2007textrunner,wu2010open,zhu2009statsnowball}. Using only lexical features will lead to problems of incoherent and uninformative
relations. Therefore, lexical features and syntactic features are used to mitigate such problems \cite{schmitz2012open, Qiu2014ZOREAS, mausam2016open}. Later, rule-based models take advantage of hand-written patterns and rules to match relations \cite{fader2011identifying,akbik2012kraken}. 
To extract relations in a fine-grained way, clause-based models determine the set of clauses and identify clause types before extracting relations \cite{del2013clausie,schmidek2014improving,angeli2015leveraging}.
\subsection{Neural Model Era} 
\label{sec:orte}


\textbf{Sequence Labeling.}
RnnOIE \cite{Stanovsky2018SupervisedOI} is the first neural method, which formulates ORTE task as a sequence labeling problem where inputs a sequence of tokens $\{x_1, x_2, ..., x_n\}$ and outputs a sequence of labels $\{l_1, l_2,..., l_n\}$ with the same length $n$ as input. RnnOIE uses a Bi-LSTM to process input features, including word embeddings, POS tags, and indicated predicates. A Softmax classifier tags a BIO label for the last layer hidden state of each token, after which relation triplets are constructed. Since one sentence usually contains more than one relation triplet, many approaches propose to avoid encoding and labeling the same input several times \cite{Kolluru2020OpenIE6IG,Bowen2021MaximalCB,Vasilkovsky2022detie}. 
SMiLe-OIE~\cite{DBLP:conf/emnlp/DongSK022} improves the model in an information-source view, using GCNs and multi-view learning to incorporate constituency and dependency information and aggregating semantic features and syntactic features by concatenating BERT and graph embeddings. SMiLe-OIE aggregates semantic and syntactic features, which many other {\openie} systems omit.

The sequence labeling paradigm is characterized by its computational efficiency, especially for large-scale text processing. It yields readily interpretable output, as each token associates itself with a specific role, such as subject, relation, object, spatial information, etc. 
It is limited by treating tokens in isolation, potentially failing to capture global context and complex relationships that extend beyond single tokens or cross sentences. Additionally, its output format may not adequately represent the nuanced variability of natural language.

\textbf{Sequence to Sequence Generation.}
\citet{Cui2018NeuralOI} casts {\openie} as a sequence-to-sequence (S2S) generation problem and proposes NeuralOIE, an encoder-decoder model generating a sequence of relation triplets conditioned by the input sentence. Facing unknown token openness problem, NeuralOIE uses the attention-based coping mechanism to enlarge the vocabulary.
IMoJIE\cite{Kolluru2020IMoJIEIM} is an iterative generative {\openie} model that uses a BERT encoder to keep encoding previous generated relation triplets and generates the next triplet with an LSTM decoder until an "EndOfExtractions" token is reached. Compared with other neural models, IMoJIE generates a variable number of diverse extractions.

The S2S paradigm excels in capturing complex relationships, as it considers the broader contextual information. It is adaptable to various languages and domains but more demand in datasets and training times. The flexible output, while better-addressing openness challenges, also poses challenges for downstream applications requiring standardized output structures.\\
\indent \textbf{Two-Stage Open Relation Extraction.}
\label{sec:orse}
Taking advantage of the remarkable representation capability of PLMs such as BERT, many researchers refine the model architecture into two stages to achieve more effective extractions.
$\textrm{Multi}^{2}\textrm{OIE}$ \cite{Ro2020Multi2OIEMO} is a two-stage labeling method. Its first stage is to label all predicates upon BERT-embedded hidden states instead of locating predicates with syntactic features. The second stage is to extract the arguments associated with each identified predicate by using a multi-head attention mechanism. In contrast to {\openie} systems that are limited to English, $\textrm{Multi}^{2}\textrm{OIE}$ is distinguished by its capability to process multilingual text. The intermediate representation can be other formats such as chunk sequence \cite{Kolluru2022AlignmentAugmentedCT} and directed acyclic \cite{yang2022open}. \\
\indent Various intermediate representations are used to enhance the pipeline's performance.
OIE@OIA \cite{Wang2022OIEOIAAA} is an adaptable {\openie} system that employs the method of Open Information expression (OIX) by parsing sentences into Open Information Annotation (OIA) Graphs. It consists of two components: an OIA generator that converts sentences into OIA graphs and a set of adaptors that trained to for versatile extraction formats. By using different intermediate representations, Chunk-OIE \cite{Dong2023OpenIE} introduces the Chunk sequence (SaC) as an intermediate representation layer while \citet{yu2022generalized} introduces directed acyclic graph (DAG) as a minimalist intermediate expression.\\
\indent \textbf{Open Relation Clustering.} 
\label{sec:orc}
The clustering-based approaches are divided into relation representation and clustering. Some studies label clusters:  
\citet{wang-etal-2022-matchprompt} and \cite{genest2022promptore} introduce an unsupervised prompt-based algorithm, MatchPrompt, which clusters sentences by leveraging representations from masked relation tokens within a prompt template. Its superb performance against traditional unsupervised methods indicates that leveraging the semantic expressive power of pre-trained models is very important.\\
\indent SelfORE \cite{hu2020selfore} propose a self-supervised learning method for learning better feature representations for clustering. SelfORE is composed of three sections: (1) encode relation instances by leveraging BERT \cite{Devlin2019BERTPO} to obtain relation representations; (2) apply adaptive clustering based on updated relation representations from (1) to assign each instance to a cluster with high confidence. In this way, pseudo labels are generated. (3) pseudo labels from (2) are used as supervision signals to train the relation classifier and update the encoder in (1). Repeat (2) until converge. Compared with other {\openie} systems, it reduces dependence on labeled data.
Based on similar self-supervised approaches, many works propose to reduce irrelevant information in relation representation \cite{Zhao2021ARC}, create pseudo labels \cite{duan-etal-2022-cluster}, and introduce human intervention during training to address the challenge of poorly clustered samples \cite{zhao-etal-2023-actively}.
During relation clustering, using complete input (sentence) representations as relation representations often leads to a significant decline in clustering performance when multiple relations exist within a single input. Semi-supervised learning has shown the best results, but the effectiveness largely depends on the quality of the annotated data.

Apart from labeled data, \textbf{knowledge bases} also benefit {\openie} by generating positive and negative instances. 
OHRE \cite{Zhang2021OpenHR} proposes a top-down hierarchy expansion algorithm to cluster and label relation instances based on the distance between the KB hierarchical structure. Existing relations are labeled with KB elements, and novel relations are labeled as children relations of existing ones.
Using a structured KB can determine the broad category of a cluster's relations, partially addressing cluster labeling issues. The KB structure can also define relation boundaries during clustering. However, errors in the KB can affect clustering accuracy, and building a high-quality KB still requires significant human effort.

\textbf{A comparison of methods in {\orte}, {\orse}, and {\orc} settings} reveals the strengths of task-specific strategies in achieving top performance. In {\orte}, MacroIE excels with the highest F1 (54.8) on CaRB, leveraging maximal clique discovery, while SpanOIE leads on OIE16 and Re-OIE16 with F1 scores of 69.4 and 77.0. In {\orse}, DragonIE tops CaRB with an F1 of 55.1, due to its efficient formulation, while OIE@OIA dominates on OIE16 and Re-OIE16 by using multi-task learning. For {\orc}, CaPL outperforms others on FewRel and TACRED through high-quality pseudo-labels, with MatchPrompt as a strong second, driven by its prompt-based framework.
\subsection{Large Language Models Era}
\label{sec:llm}
The recent evolution and emergence of Large Language Models (LLMs), such as GPT-4 \cite{openai2024gpt4}, ChatGPT \cite{openai2023chatgpt}, and Llama 2 \cite{touvron2023llama}, have significantly advanced the field of NLP. Their remarkable capabilities in text understanding, generation, and generalization have led to a surge of interest in generative IE methods \cite{qi-etal-2023-preserving, xu2023large}. Recent studies have employed LLMs for {\openie} tasks by transforming input text through specific instructions or schemas. This approach facilitates tasks such as triplet extraction and relation classification under the structured language generation framework. It allows for a versatile task configuration where diverse forms of input text can be processed to generate structured relational triplets uniformly.\\
\indent\textbf{Zero-Shot.} \citet{wang-etal-2022-ielm} propose IELM, a benchmark for assessing the zero-shot performance of GPT-2 \cite{radford2019language} by encoding entity pairs in the input and extracting relations associated with each entity pair. On large-scale evaluation on various {\openie} benchmark tasks, research has shown that the zero-shot performance of leading LLMs, such as ChatGPT, still falls short of the state-of-the-art supervised methods \cite{han2023information, qi-etal-2023-preserving}, specifically on more challenging tasks \cite{li2023evaluating}. This shortfall is partly because LLMs struggle to distinguish irrelevant context from long-tail target types and relevant relations \cite{ling2023improving, han2023information}.\\
\indent \textbf{Fine-Tuning and Few-Shot.} Consequently, efforts have been made to fine-tune pre-trained LLMs or employ in-context learning prompting strategies to utilize and enhance the instruction-following ability of LLMs. For example, \citet{lu2023pivoine} addresses open-world information extraction, including unrestricted entity and relation detection, as an instruction-following generative task, and develops PIVOINE, refined via the process of instruction tuning, demonstrates the capacity to accurately extract pertinent entity information in accordance with directives articulated in natural language. Such a capability endows it with the flexibility to adapt to various information extraction tasks and requirements. To minimize the need for extensive fine-tuning of LLMs, \citet{ling2023improving} proposes various in-context learning strategies for performing relation triplet generation to improve the instruction-following ability of LLMs, and introduces an uncertainty quantification module to increase the confidence in the generated answers. \citet{qi2023mastering} proposes to construct a consistent reasoning environment by mitigating the distributional discrepancy between test samples and LLMs. This strategy aims to improve the few-shot reasoning capability of LLMs on specific {\openie} tasks.

\section{Discussion} 
\label{sec:discussion}
This section reviews the diverse sources of information used by {\openie} models and discusses current limitations and future prospects, offering a comprehensive overview of the field's evolving trajectory.

\subsection{Co-Evolution of {\openie}: Task Settings and Model Capabilities}
\label{sec:discuss_coevolution}


In this section, we unveil the connection between task settings and model capabilities in handling various features and information, demonstrating the intertwined development of both aspects. 

\textbf{Input-based information} refers to features explicitly or implicitly present in the input unstructured text. Early {\openie} models extensively utilized explicit information such as \textbf{\textit{shallow syntactic information}}, including part of speech (POS) tags and noun-phrase (NP) chunks \cite{banko2007open, wu2010open, fader2011identifying}. This approach is reliable, yet it does not capture all relation types \cite{Stanovsky2018SupervisedOI}, leading to the increasing use of \textbf{\textit{deep dependency information}}, which reveals word dependencies within sentences \cite{vo2018open, elsahar2017unsupervised}. Subsequent {\openie} models have emphasized the use of \textbf{\textit{semantic information}} to grasp literal meanings and linguistic structures, thereby enhancing the expression of relations despite the risk of over-specificity \cite{Vashishth2018CESICO, Wu2018TowardsPO}. Recent models, including pre-trained language models, combine syntactic and semantic information to improve accuracy \cite{Hwang2020BERTbasedKO, ni2021explore}. Further details in Appendix \ref{subsec:appendix_source_input}.

\textbf{External information} supplements {\openie} systems to enhance model performance. Early systems employ \textbf{\textit{expert rules}}, including heuristic rules that integrate domain knowledge and assist in error tracing and resolution, based on syntactic analyses like POS-tagging \cite{chiticariu-etal-2013-rule, fader2011identifying}. Following this, the integration of \textbf{\textit{hierarchical information}} from knowledge bases (KBs) advances knowledge representation learning. This integration provides structured hierarchies and detailed factual knowledge, supporting more organized relation extraction and data augmentation \cite{Xie2016RepresentationLO, Zhang2021OpenHR, Fangchao2021ElementIF}. With the developments of LLMs recently, the \textbf{\textit{pre-trained knowledge}} within these models is utilized, encapsulating extensive relational data \cite{jiang2020can, petroni2020context} and enabling efficient retrieval with well-designed instructions. The strong generalization capabilities of LLM-based approaches allow them to embrace \textbf{\textit{open-world knowledge}}, making them more robust and adaptable to various tasks and real-world applications. Further details in Appendix \ref{subsec:appendix_source_external}. 



\subsection{Transforming OpenIE: The Impact of LLMs} 
\label{subsec:discuss_llm}

When comparing the \textbf{performance of LLMs with pre-LLM approaches}, we see that LLMs have significantly advanced the task of OpenIE, often outperforming traditional methods. Zero-shot LLMs have achieved impressive and state-of-the-art (SOTA) results in various scenarios when evaluated on classic metrics such as token-level scorers \cite{li2023evaluating, wang-etal-2022-ielm}. However, these models struggle with long-tail and more challenging tasks \cite{gao2023exploring}. A major challenge for LLMs, compared to pre-LLM approaches like sequence tagging, is the issue of hallucination, which frequently occurs in various natural language generation tasks \cite{ji2023survey}, making faithfulness and reliability significant concerns. Traditional generative-based approaches from the pre-LLM era also suffer from errors such as redundant and incorrect extractions \cite{schneider-etal-2017-analysing, zhou2022survey}, known as intrinsic hallucination. In contrast, LLM-based methods face the risk of both intrinsic hallucination and generating information unsupported by the original context or additional references, known as extrinsic hallucination \cite{zhu2023large, DBLP:journals/corr/abs-2307-11019, li2023evaluating}. Despite these challenges, few-shot learning and fine-tuning can help mitigate issues related to long-tail challenges and hallucination to some extent. Additionally, until fundamental improvements in LLMs fully address these shortcomings, incorporating traditional approaches as supplementary supervisors when using LLM-based methods could potentially enhance reliability.

We also observe trends in developing \textbf{universal paradigms for tackling various IE tasks.} Recent advancements and the robust generalization capabilities of LLMs have led to the exploration of universal frameworks designed to address all IE tasks (UIE). These frameworks aim to leverage the shared capabilities inherent in IE, while also uncovering and learning from the dependencies between various tasks \cite{xu2023large}. This approach marks a significant shift from focusing on isolated subtasks, such as OpenIE, to a more integrated methodology that seeks a comprehensive understanding of the domain. The prevailing trajectory in developing universal IE frameworks is to establish unified, structured schemas, either natural language-based \cite{wang-etal-2022-deepstruct, lu-etal-2022-unified, 10.1609/aaai.v37i11.26563} or code-based \cite{li-etal-2023-codeie, guo2023retrieval, sainz2023gollie}, to transform various IE tasks into a uniform task of structural information extraction while preserving the flexibility to adapt to the unique aspects of different tasks. More details on these approaches are provided in Appendix \ref{sec:llmAPP}.

\textit{Is {\openie} research going to its ends?} \textbf{LLMs bridge the gap between standard IE and OpenIE.} LLMs are naturally suited for OpenIE, even under zero-shot scenarios, as they address both standard IE and OpenIE within the same task setting. In this setting, schemas and templates are designed to extract desired structural information. The primary difference is that standard IE schemas include more restrictions to limit the set of relations and entities. The flexibility and strong performance of LLMs in tackling various IE tasks through zero-shot and few-shot prompting, without requiring model updates, is attributed to their robust generalization ability acquired through pre-training. With this generalization capability, addressing both standard IE and OpenIE may not require fundamentally different methods; the main distinction lies in schema design. This significantly blurs the boundaries between standard IE and OpenIE. In the future, OpenIE might be viewed as a more complex and challenging scenario within IE tasks, rather than being distinctly separate from standard IE. Though we refrain from making a definitive conclusion, we can foresee OpenIE potentially merging into the broader scope of standard IE.
\subsection{Future Directions} 
\label{sec:futuredir}

Although we see the momentum of blurred gaps between OpenIE and standard IE with the impact of LLM, the fundamental task itself remains. Then \textit{how can traditional {\openie} research inspire IE research in the LLM era?} Following we discuss future directions draw from reflections on a chronological perspective.

\textbf{OpenIE datasets} are growing but remain small and narrow in scope. Insights from traditional {\openie} research suggest that future expansions are needed to include more languages, domains, and broader sources. LLMs offer the opportunity to improve this through their capabilities in synthesizing and augmenting data. While synthesized datasets have been extensively explored within the domain of standard IE \cite{zhang-etal-2023-llmaaa, xu-etal-2023-s2ynre}, with researchers claiming that the proposed methods can be adapted for OpenIE \cite{josifoski-etal-2023-exploiting}, there is a notable gap regarding comprehensive studies on synthesized datasets for OpenIE. Addressing this gap could facilitate the creation of cross-domain datasets and the integration of existing datasets and tasks.

As discussed in Section \ref{subsec:discuss_llm}, LLMs enable the exploration of various IE tasks with universal frameworks (UIE). Despite advances, most LLM-based UIE systems focus on standard IE tasks and often overlook OpenIE, a complex challenge within the IE spectrum. LLMs are inherently suited for OpenIE due to their extensive pre-trained knowledge. 
Therefore, the \textbf{primary challenge of LLMs} lies not in extracting relational information but in accurately interpreting and following task-specific instructions, as well as mitigating hallucination. Integrating traditional approaches into LLM-based frameworks might address these current shortcomings of LLMs. Additionally, these approaches can provide insights for developing more robust, faithful, and reliable fundamental LLMs.

\textbf{More comprehensive automatic metrics} are needed to evaluate LLM-based approaches. As discussed in Sections \ref{sec:tasksettings} and \ref{sec:eva}, task settings and corresponding evaluation metrics develop hand-in-hand. Now the changes brought by LLM calls for a more holistic and update-to-date evaluation metrics in many NLP tasks \cite{gong2024cream, ai2024novascore, ai2024defending, ai2024enhancing}. Current efforts explore different options, as noted in Section \ref{sec:eva} and Appendix \ref{appendix:evaluation}, but aspects such as faithfulness still rely heavily on human evaluation and lack a commonly accepted metric. Developing new, comprehensive automatic evaluation methods that capture nuanced aspects of OpenIE output, such as semantic coherence, factual accuracy, and information completeness, will lead to more robust and reliable LLM-based OpenIE systems. These metrics can address the unique challenges posed by LLMs, including their propensity for generating diverse and open-ended outputs.

\textbf{Multilinguality} is critical, yet there is a noticeable gap in multilingual capabilities. 
Although the models discussed in Appendix \ref{sec:other} show efforts to adapt across languages, there is a notable deficiency in initiatives using LLMs to tackle these challenges. This highlights the need to develop datasets, metrics, and robust benchmarks, particularly for low-resource languages, to enhance the capabilities of multilingual OpenIE.



\textbf{Latency, cost, and distillation}. Reviewing the development of models for OpenIE, we see the trend that recent development in LLMs introduce a more expensive system with higher latency, especially using close sourced LLMs such as GPT. Although the rapid iteration of models shows cuts on cost and latency, more effective solutions could be possible with knowledge distillation from LLMs onto specialized SLMs, revisiting the prior neural model era we discussed.


\section{Conclusion}
\label{sec:conclusion}

We systematically survey the development of OpenIE from a chronological perspective, highlighting historical trends in task settings and model development. We draw important connections and derive lessons from the influence of technology on task settings, examining the advantages and disadvantages of both past and present methods. Furthermore, we explore the increasingly blurred distinctions between OpenIE and standard IE. For researchers in LLMs, past work should not be overlooked; instead, it should serve as a valuable resource for future inquiries.



\newpage
\section*{Limitations}
\label{sec:limitations}


Our survey primarily concentrates on the chronological evolution of OpenIE technologies and their alignment with significant milestones in NLP development. Consequently, we have not  covered multi-domain and multi-lingual datasets or methodologies extensively. While we do address some non-English datasets, specifically Mandarin, and briefly mention multilingual models in Appendix \ref{sec:methodAPP} and model applications across various domains in Appendix \ref{subsec:domain}, these discussions are not the focal point of our analysis. This limitation is intentional in order to maintain a clear focus on the historical progression of the field rather than the breadth of dataset diversity or the adaptability of methodologies across languages and domains.

Another potential limitation is our survey's emphasis on the macro aspects of the OpenIE field rather than detailed, micro-level analysis of specific methodologies. As outlined in Section \ref{sec:intro}, many existing surveys already cover methodologies and models from the pre-LLM era, and we felt that redundant elaboration on these would not add significant value. Post-LLM, despite substantial research leveraging LLMs for standard IE tasks, there is still a scarcity of studies specifically applying LLMs to OpenIE tasks. This scarcity has constrained our ability to conduct an in-depth survey focused exclusively on LLM methodologies within OpenIE. Nonetheless, from the existing work on LLMs in standard IE and UIE, detailed in Appendix \ref{sec:llmAPP}, we observe emerging trends that warrant a macro-level analysis. Our approach of integrating and reviewing the field through a historical lens is essential to provide a comprehensive view, enabling a clearer understanding of the task and aiding in the development of a more defined future roadmap.
\section*{Acknowledgements}
The work was supported in part by Alibaba Innovative Research (AIR) project support funding. We thank Yafu Li and all reviewers for their generous help and advice during this research. 
This work was supported in part by the US Department of Defense under the DARPA CCU program. Any opinions expressed herein are those of the authors and do not necessarily reflect the views of the U.S. Department of Defense or the U.S. Government.

\bibliography{custom, llm-references, pre_neural}
\clearpage
\appendix

\section{Open IE Methodologies in Details}
\label{sec:methodAPP}

A Chronological Overview of Open IE methods are summarized in Figure \ref{fig:method_chron_view}.


\subsection{Open Relation Triplet Extraction}
\subsubsection{Labeling}
OpenIE6 \cite{Kolluru2020OpenIE6IG} adopts a novel Iterative Grid Labeling (IGL) architecture, with which OpenIE is modeled as a 2-D grid labeling problem. Each extraction corresponds to one row in the grid. Iterative assignments of labels assist the model in capturing dependencies among extractions without re-encoding.

Owing to the outstanding performance of PLMs, many researchers extend the sequence labeling task to other problems.
MacroIE\cite{Bowen2021MaximalCB} reformulates the OpenIE as a non-parametric process of finding maximal cliques from the graph. It uses a non-autoregressive framework to mitigate the issue of enforced order and error accumulation during extraction. 
DetIE \cite{Vasilkovsky2022detie} casts the task to a direct set prediction problem. This encoder-only model extracts a predefined number of possible triplets (proposals) by generating multiple labeled sequences in parallel, and its order-agnostic loss based on bipartite matching ensures the predictions are unique. 
\subsection{Open Relation Span Extraction}
 \textsc{Gen2OIE} \cite{Kolluru2022AlignmentAugmentedCT} extends to a generative paradigm operating in two stages. It first generates all possible relations from input sentences. Then, it produces extractions for each generated relation. This generative approach allows for overlapping relations and multiple extractions with the same relation.
 
 \citet{jia2022hybrid} propose a hybrid neural network model (HNN4ORT) for open relation tagging. The model employs the Ordered Neurons LSTM \cite{Shen2019OrderedNI} to encode potential syntactic information for capturing associations among arguments and relations. It also adopts a novel Dual Aware Mechanism, integrating Local-aware Attention and Global-aware Convolution.
 QuORE \cite{yang2022open} is a framework to extract single/multi-span relations and detect non-existent relationships, given an argument tuple and its context. The model uses a manually defined template to map the argument tuple into a query. It concatenates and encodes the query together with the context to generate sequence embedding, with which this framework dynamically determines a sub-module (Single-span Extraction or Query-based Sequence Labeling) to label the potential relation(s) in the context.

 Inspired by OIA, Chunk-OIE \cite{Dong2023OpenIE} introduces the concept of Sentence as Chunk sequence (SaC) as an intermediate representation layer, utilizing chunking to divide sentences into related non-overlapping phrases. 
\citet{yu2022generalized} introduce directed acyclic graph (DAG) as a minimalist expression of open fact in order to reduce the extraction complexity and improves the generalization behavior. They propose DragonIE which leverages the sequential priors to reduce the complexity of function space (edge number and type) in the previous graph-based model from quadratic to linear, while avoiding auto-regressive extraction in sequence-based models.
\subsection{Open Relation Clustering}
\citet{Lechevrel2017CombiningSA} select core dependency phrases to capture the semantics of the relations between entities. The design rules are based on the length of the dependency phrase in the dependency path, which sometimes contains more than one dependency phrase that uses all terms and brings in irrelevant information. Each relation instance is clustered on the basis of the semantics of core dependency phrases. Finally, clusters are named by the core dependency phrase most similar to the center vector of the cluster.

Instead of directly cutting less irrelevant information, \citet{elsahar2017unsupervised} propose a more resilient approach based on the shortest dependency path. The model generates representations of relation instances by assigning a higher weight to word embedding of terms in the dependency path and then reduces feature dimensions by PCA \cite{Shen2009PrincipalCA}. Although the model ignores noisy terms in the dependency path, re-weighting is a forward-looking idea resembling the subsequent attention mechanism. 

The key idea of \citet{Fangchao2021ElementIF} is based on blocking backdoor paths from a causal view \cite{Pearl2000CausalityMR}. The intervened context is generated by a generative PLM, while entities are intervened by placing them with three-level hierarchical entities in KB. Model parameters are optimized by those intervened instances via contrastive learning. The learned model encodes each instance into its representations, before using clustering algorithms.
\subsection{Neural Model Era: Other Settings} \label{sec:other}
\textbf{Translation.}
\citet{wang2021zero} cast information extraction tasks into a text-to-triplet translation problem. They introduce \textsc{DeepEx}, a framework that translates NP-chunked sentences to relational triplets in a zero-shot setting. This translation process consists of two steps: generating a set of candidate triplets and ranking them. 

\noindent\textbf{Multilingual.}
\textsc{milIE} \cite{Kotnis2022MILIEM} is an integrated model of a rule-based system and a neural system, which extracts triplet slots iteratively from simple to complex, conditioning on preceding extractions. The iterative nature guarantees the model to perform well in a multilingual setting. $\textrm{Multi}^{2}\textrm{OIE}$ \cite{Ro2020Multi2OIEMO} also has a multilingual version based on multilingual-BERT, which makes it able to deal with various languages. Differently, LOREM \cite{Harting2020LOREMLO} trains two types of models, language-individual models, and language-consistent models and incorporates multilingual, aligned word embeddings to enhance model performance.

\section{LLMs for IE in general}
\label{sec:llmAPP}

In Section \ref{sec:llm}, we begin by reviewing the work that utilizes LLMs to address OpenIE. Here, we 1). broaden our scope to introduce some emerging trends and paradigms in universal information extraction. For an in-depth exploration of how LLMs are applied to closed relation extraction and other IE tasks, we refer readers to the survey by \citet{xu2023large} for comprehensive details. Moreover, we 2). further expand our discussion to explore research that integrates LLMs into IE system pipelines, beyond merely using them for direct IE task solution. We 3). also includes an discussion of current trends in IE dataset using LLMs that shed light on the future of datasets on openIE. 

We believe this broader perspective provides readers with a comprehensive understanding of current trends and future directions in OpenIE and generic IE in the LLM era, enhancing their grasp of the field's evolving dynamics.

\subsection{Universal Information Extraction}
Recent advancements and the robust generalization capabilities of LLMs have led to the exploration of universal frameworks designed to tackle all IE tasks (UIE). These frameworks aim to harness the shared capabilities inherent in IE, while also uncovering and learning from the dependencies that exist between various tasks \cite{xu2023large}. This approach marks a significant shift from focusing on isolated subtasks such as OpenIE to a more integrated methodology that seeks to understand a more integrated and comprehensive understanding of the domain. 

\textbf{Natural Language-Based Schema.} A prevailing trend in developing universal IE frameworks is to establish a unified, structured natural language schema for diverse subtasks, designed for schema-prompting LLMs. For instance, \citet{wang-etal-2022-deepstruct} introduce DeepStruct, which reformulates various IE tasks as triplet generation tasks, using generalized task-specific prefixes in prompts and pretraining LLMs to comprehend text structures. \citet{lu-etal-2022-unified} propose UIE, encoding different extraction structures uniformly through a structured extraction language and adaptively generating specific extractions with a schema-based prompt strategy. Similarly, \citet{10.1609/aaai.v37i11.26563} present USM, encoding different schemas and input texts together to enable structuring and conceptualizing, aiming for a single model that addresses all tasks. Building on UIE and USM, \citet{wang2023instructuie} introduce InstructUIE, which models various IE tasks uniformly with descriptive natural language instructions for instruction tuning, exploiting inter-task dependencies.

\textbf{Code-Based Schema.} Despite their empirical success, natural language-based approaches face challenges in generating outputs for IE tasks due to the distinct syntax and structure that differ from the training data of LLMs \cite{10.1145/3641850}. In response to these limitations and leveraging recent advancements in Code-LLMs \cite{chen2021evaluating}, researchers have begun to utilize Code-LLMs for structure generation tasks \cite{wang2022code4struct}, as code, a formalized language, adeptly describes structural knowledge across various schemas universally \cite{guo2023retrieval}. For instance, \citet{li-etal-2023-codeie} present CodeIE, which translates structured prediction tasks such as NER and RE into code generation, employing Python functions to create task-specific schemas and using few-shot learning to instruct Code-LLMs. \citet{guo2023retrieval} introduce Code4UIE, utilizing Python classes to define task-specific schemas for diverse structural knowledge universally. Similarly, \citet{sainz2023gollie} propose GoLLIE, which employs Python classes to encode IE tasks and, in addition, integrates task-specific guidelines as docstrings, enhancing the robustness of fine-tuned Code-LLMs to schemas not encountered during training.

\subsection{Role of LLMs in IE System}
In addition to directly addressing IE tasks, LLMs have shown utility as specific components within IE system pipelines, including data synthesis for IE model training and knowledge retrieval for downstream IE tasks.

\textbf{Data Synthesis.} A prominent application of LLMs in IE systems is the synthesis of high-quality training data, as data curation through human annotation is time-consuming and labor-intensive. One approach employs LLMs as annotators within a learning loop \cite{zhang2023llmaaa}, while another strategy involves using LLMs to inversely generate natural language text from structured data inputs \cite{josifoski-etal-2023-exploiting, ma2023star}, thereby producing large-scale, high-quality training data for IE tasks.

\textbf{Knowledge Retrieval.} Another research direction exploits the capability of LLMs, developed through pre-training, as implicit knowledge bases to generate or retrieve relevant context for downstream IE tasks. For instance, \citet{li2023prompting, li2024llms} employ LLMs to generate auxiliary knowledge improving multimodal IE tasks. \citet{amalvy2023learning} demonstrate that pre-trained LLMs possess inherent knowledge of the datasets they work on, and use these models to generate a context retrieval dataset, enhancing NER performance on long documents.

\subsection{IE in Different Domains} 
\label{subsec:domain}
The development of Information Extraction (IE) has seen significant advancements across various domains, including Multimodal IE, Medical Information Extraction, and the application of Code Models for IE tasks. These developments have been particularly enhanced by the integration of Large Language Models (LLMs), which have improved downstream task performance through their use in model architecture and as tools for annotation and training guidance.

\textbf{Medical Information Extraction} has greatly benefited from the use of LLMs as efficient tools for annotation, as highlighted in research by \citet{pmlr-v225-goel23a, meoni-etal-2023-large}. These applications enhance data quality and contribute to the overall improvement of model performance.

\textbf{Multimodal IE}  tasks, such as Multimodal Named Entity Recognition (MNER) and Multimodal Relation Extraction (MRE), have advanced through frameworks that capitalize on the capabilities of LLMs in IE. \citet{cai-etal-2023-context} propsed to use in-context learning (ICL) ability in ChatGPT to help Few-Shot MNER by employing in-context learning to convert visual data into text and select relevant examples for effective entity recognition. \citet{li-etal-2023-prompting} tackles MNER on social media by efficient usage of generated knowledge and improved generalization, which utilizes ChatGPT as an implicit knowledge base for generating auxiliary knowledge to aid entity prediction. \citet{chen2023chainofthought} distill the reasoning ability of LLMs by using "chain of thought" (CoT) to elicit reasoning capability from LLMs across multiple dimensions to improve MNER and MRE.

\textbf{Code generative LLMs} have found application in performing IE tasks such as Universal Information Extraction (UIE) \cite{li-etal-2023-codeie, guo2023retrievalaugmented}, Event Structure Prediction \cite{wang-etal-2023-code4struct}, and Generative Knowledge Graph \cite{10.1145/3641850}, where researchers convert the structured output in the form of code instead of natural language, and utilize generative LLMs of code (Code-LLMs) by designing code-style prompts and formulating these IE tasks as code generation tasks.

Leveraging LLMs across different domains has not only broadened the scope of IE applications but also significantly improved the effectiveness and efficiency of extraction tasks.

\section{Datasets}
\label{appendix:datasets}

\textbf{Question Answering (QA) derived datasets} are converted from other crowdsourced QA datasets. 
OIE2016 \cite{Stanovsky2016CreatingAL} is one of the most popular OpenIE benchmarks, which leverages QA-SRL \cite{He2015QuestionAnswerDS} annotations.
AW-OIE \cite{Stanovsky2018SupervisedOI} extends the OIE2016 training set with extractions from QAMR dataset \cite{Michael2017CrowdsourcingQM}.
The OIE2016 and AW-OIE datasets are the first datasets used for supervised OpenIE. However, because of its coarse-grained generation method, OIE2016 has some problematic annotations and extractions. 
On the basis of OIE2016, Re-OIE2016 \cite{Zhan2020SpanMF} and CaRB \cite{Bhardwaj2019CaRBAC} re-annotate part of the dataset.
LSOIE \cite{Solawetz2021LSOIEAL} is created by converting QA-SRL 2.0 dataset \cite{FitzGerald2018LargeScaleQP} to a large-scale OpenIE dataset, which claims 20 times larger than the next largest human-annotated OpenIE dataset.

\textbf{Crowdsourced datasets} are created from direct human annotation, including WiRe57 \cite{Lchelle2019WiRe57A}, SAOKE dataset \cite{Sun2018LogicianAU}, and BenchIE dataset \cite{Gashteovski2021BenchIEOI}. WiRe57 is created based on a small corpus containing 57 sentences from 5 documents by two annotators following a pipeline. SAOKE dataset is generated from Baidu Baike, a free online Chinese encyclopedia, like Wikipedia, containing a single/multi-span relation and binary/polyadic arguments in a tuple. It is built in a predefined format, which assures its completeness, accurateness, atomicity, and compactness.

\textbf{Knowledge Base (KB) derived datasets} are established by aligning triplets in KBs with text in the corpus. 
Several works \cite{Mintz2009DistantSF,Yao2011StructuredRD} have aligned the New York Times corpus \cite{sandhaus2008new} with Freebase \cite{Bollacker2008FreebaseAC} triplets, resulting in several variations of the same dataset, NYT-FB.
FewRel \cite{Han2018FewRelAL} is created by aligning relations of given entity pairs in Wikipedia sentences with distant supervision, and then filtered by human annotators.
\citet{ElSahar2018TRExAL} propose a pipeline to align Wikipedia corpus with Wikidata \cite{Vrandei2012WikidataAN} and generate T-REx. By filtering triplets and selecting sentences, \citet{hu2020selfore} create T-REx SPO and T-REx DS. In addition, COER \cite{Jia2018ChineseOR}, a large-scale Chinese knowledge base dataset, is automatically created by an unsupervised open extractor from diverse and heterogeneous web text, including encyclopedia and news. Overall, KB derived datasets are mostly used in open relation clustering task setting, illustrated in Section~\ref{sec:orc}, whereas QA derived and crowdsourced datasets are usually used in open relational triplet extraction (Section~\ref{sec:orte}) and open relation span extraction task settings (Section~\ref{sec:orse}).

\textbf{Instruction-based datasets} transform IE tasks into tasks requiring instruction-following, thus harnessing the capabilities of LLMs. One strategy involves integrating various existing IE datasets into a unified-format benchmark dataset with specifically designed instructions \cite{wang2023instructuie, lu-etal-2022-unified}. Alternatively, instruction-based IE datasets such as \textsc{InstructOpenWiki} \cite{lu2023pivoine} and \textsc{InstructIE} \cite{gui2023instructie}, or structured IE datasets like Wikidata-OIE \cite{wang-etal-2022-ielm}—derived from Wikidata and Wikipedia—are created. The first method primarily focuses on ClosedIE tasks, while the second offers more flexibility in generating OpenIE datasets \cite{lu2023pivoine, wang-etal-2022-ielm}.

\label{appendix:llm_datasets}

\textbf{Synthesized datasets using LLMs} on IE expands significantly compared to previous ones in both the size of the datasets and data qualities. While the methodologies for synthesizing these datasets have been extensively explored within the domain of closed Information Extraction (ClosedIE) \cite{zhang-etal-2023-llmaaa, xu-etal-2023-s2ynre}, where researchers claims the proposed methods can be adapted for OpenIE setting \cite{josifoski-etal-2023-exploiting}, there remains a notable gap in the literature regarding comprehensive studies on synthesized datasets for OpenIE.

\section{Evaluation}
\label{appendix:evaluation}
\textbf{Token-level Scorers.}
To allow some flexibility (e.g., omissions of prepositions or auxiliaries), if automated extraction of the model and the gold triplet agree on the grammatical head of all of their elements (predicate and arguments), OIE2016 \cite{Stanovsky2016CreatingAL} takes it as matched. \citet{Lchelle2019WiRe57A} penalize the verbosity of automated extractions as well as the omission of parts of a gold triplet by computing precision and recall at token-level in WiRe57. Their precision is the proportion of extracted words that are found in the gold triplet, while recall is the proportion of reference words found in extractions. To improve token-level scorers, CaRB \cite{Bhardwaj2019CaRBAC} computes precision and recall pairwise by creating an all-pair matching table, with each column as extracted triplet and each row as gold triplet. When assessing LLM extracted spans, \citet{han2023information} report the ratio of invalid responses, which include incorrect formats and content not aligned with task-specific prompts. As generative models, LLMs aim to mimic human-like responses and often generate longer text than the gold standard annotations.

\textbf{Fact-level Scorers.} SAOKE \cite{Sun2018LogicianAU} measures to what extent gold triplets and extracted triplets imply the same facts and then calculates precision and recall. BenchIE \cite{Gashteovski2021BenchIEOI} introduces \textit{fact synset}: a set of all possible extractions (i.e., different surface forms) for a given fact type (e.g., VP-mediated facts) that are instances of the same fact. It takes the informational equivalence of extractions into account by exactly matching extracted triplets with the gold fact synsets. In assessing outputs from LLMs, \citet{li2023evaluating} have ChatGPT provide justifications for its predictions and use domain expert annotation to verify their faithfulness relative to the input.

\subsection{Evaluation Metrics of ORC}

\subsection*{\boldmath $B^3$ Metric}

$B^3$ is an instance-based method that computes the precision and recall for each instance in the test set by comparing the cluster containing the instance in the prediction results with the cluster containing the instance in the ground truth (golden answer). $B^3$ averages the precision and recall for each instance and then computes the harmonic mean to provide a final score.

\textbf{Precision for an element $i$:}

\[
P(i) = \frac{|C_i \cap T_i|}{|C_i|}
\]

Where:
$C_i$ is the set of elements in the same cluster as $i$. $T_i$ is the set of elements in the same true class as $i$.

\textbf{Recall for an element $i$:}

\[
R(i) = \frac{|C_i \cap T_i|}{|T_i|}
\]

\textbf{Average Precision and Recall:}

\[
P = \frac{1}{N} \sum_{i=1}^{N} P(i), \quad R = \frac{1}{N} \sum_{i=1}^{N} R(i)
\]

Where $N$ is the total number of elements.

\textbf{F1 Score:}

\[
F1 = 2 \cdot \frac{P \cdot R}{P + R}
\]

\subsection*{V-measure}

V-measure is an entropy-based, instance-based clustering evaluation metric that introduces conditional entropy, requiring higher purity of clusters. Compared to $B^3$, the existence of a few incorrect instances in a relatively pure cluster decreases the score more significantly, thereby punishing clustering results more harshly. The V-measure F1 score is the harmonic mean of homogeneity and completeness.

\textbf{Homogeneity $h$:}

\[
h = 1 - \frac{H(C \mid K)}{H(C)}
\]

Where $H(C \mid K)$ is the conditional entropy of the classes given the cluster assignments; $H(C)$ is the entropy of the classes.

\textbf{Completeness $c$:}

\[
c = 1 - \frac{H(K \mid C)}{H(K)}
\]

Where $H(K \mid C)$ is the conditional entropy of the clusters given the class assignments; $H(K)$ is the entropy of the clusters.

\textbf{V-measure:}

\[
V = \frac{2 \cdot h \cdot c}{h + c}
\]

\subsection*{Adjusted Rand Index (ARI)}

Adjusted Rand Index (ARI) measures the similarity between predicted and true clusterings by counting all pair-wise assignments in the same or different groups. It adjusts for chance groupings, with a threshold ranging from 0 to 1, representing random groupings to perfectly accurate groupings. Compared to $B^3$ and V-measure, ARI is less sensitive to extreme values like precision or homogeneity, providing a more balanced evaluation.

\textbf{Rand Index (RI):}

\[
RI = \frac{a + b}{a + b + c + d}
\]

Where $a$ is the number of pairs of elements that are in the same set in both the predicted and true clusterings, $b$ is the number of pairs of elements in different sets in both the predicted and true clusterings, $c$ is the number of pairs of elements in the same set in the true clustering but not in the predicted clustering, and $d$ is the number of pairs of elements in the same set in the predicted clustering but not in the true clustering.

\textbf{Adjusted Rand Index (ARI):}

\[
\scalebox{1}{
$ARI = \frac{\sum_{ij} \binom{n_{ij}}{2} - \left[ \sum_i \binom{a_i}{2} \sum_j \binom{b_j}{2} \right] / \binom{n}{2}}{\frac{1}{2} \left[ \sum_i \binom{a_i}{2} + \sum_j \binom{b_j}{2} \right] - \left[ \sum_i \binom{a_i}{2} \sum_j \binom{b_j}{2} \right] / \binom{n}{2}}$
}
\]

Where $n_{ij}$ is the number of elements in both predicted cluster $i$ and true cluster $j$.

\section{Source of Information}

Section \ref{sec:discuss_coevolution} provides a brief overview of the sources of information utilized in OpenIE models. This section offers a detailed discussion of each specific information source.

\subsection{Input-based Information}
\label{subsec:appendix_source_input}

\textbf{Shallow syntactic information}\label{shallowsyn} such as part of speech (POS) tags and noun-phrase (NP) chunks abstract input sentences into patterns. It is pervasively used in the early work of OpenIE as an essential model feature \cite{banko2007open, wu2010open, fader2011identifying}. In rule-based models, those patterns directly determine whether the input text contains certain relations or not \cite{Xavier2013OpenIE,A2013ComparisonOO}. Shallow syntactic information is reliable because there is a clear relationship between the relation type and the syntactic information in English \cite{banko2007open}. However, merely using shallow syntactic information can not discover all relation types. Subsequent work uses shallow syntactic information as part of the input and incorporates additional features to enhance the model performance \cite{Stanovsky2018SupervisedOI}.

\textbf{Deep dependency information} shows the dependency between words in a sentence, which can be used directly to find relations \cite{vo2018open}. But because dependency analysis is more complex and time-consuming than shallow syntactic analysis, such information source was not popular in early OpenIE studies. It was the second generation of OpenIE models that brought dependency parsing to great attention. Right now, dependency information is still used as part of the model input, though with less popularity and sometimes not directly. \citet{elsahar2017unsupervised} make use of the dependency path to give higher weight to words between two named entities, in which way the model only uses dependency information as a supplement and relies more on the semantic meaning to extract information.


    
\textbf{Semantic information} captures not only linguistic structures of sentences but literal meanings of phrases, which can express more diverse and fitting relations compared to syntactic patterns. However, semantic information can also be too specific and hence lead to the canonicalizing problem \citep{Galrraga2014CanonicalizingOK,Vashishth2018CESICO,Wu2018TowardsPO}.
The second generation of OpenIE models has tried to use semantic information via semantic role labeling, for example \textsc{Examplar} \cite{mesquita2013effectiveness}, or via dependency parsing, for instance OLLIE \cite{schmitz2012open}. There were also attempts to use WordNet output to comprise semantic information \cite{liu2012joint}. The third generation of OpenIE models typically use the word and sentence representations obtained from pre-trained language models \cite{Kolluru2020IMoJIEIM, Hwang2020BERTbasedKO, Xinwei2020OpenIE}. These representations contain both syntactic and semantic information \cite{jawahar2019does}. Meanwhile, some OpenIE models use word embeddings from word embedders such as GloVe, ELMo, and Word2Vec to capture semantic information \cite{ni2021explore}.

\subsection{External Knowledge}
\label{subsec:appendix_source_external}

\textbf{Expert rules} are knowledge imported in the form of heuristic rules. It is easy for rule-based OpenIE systems to incorporate domain knowledge as well as to trace and fix errors \cite{chiticariu-etal-2013-rule}. Heuristic rules can be employed to avoid incoherent extractions \cite{fader2011identifying}.
For example, verb words between two entities are likely to be the relation. Thus, to alleviate incoherence, a rule can be defined: \emph{If there are multiple possible matches for a single verb, the shortest possible match is chosen.} Based on patterns generated from POS-tagging, dependency parse, and other syntactic analyses, different rules can be created.  

\textbf{Hierarchical information} that implicitly exists in languages, which can be explicitly exhibited by knowledge bases, benefits knowledge representation learning \cite{Wang2014KnowledgeGA, Lin2015LearningEA,Hu2015EntityHE, Xie2016RepresentationLO}. In addition, KBs contain fine-grained factual knowledge that provides background information and hierarchical structures needed for relation extraction. Compared to traditional clustering, KB can provide hierarchical information that helps represent and cluster relations in a more organized way \cite{Zhang2021OpenHR} and hierarchical factual knowledge for data augmentation \cite{Fangchao2021ElementIF}. \\

\textbf{Pre-trained knowledge} of language models, particularly LLMs, exhibit substantial potential to encapsulate relational knowledge \cite{jiang2020can, petroni2020context}. Unlike smaller models, which require learning from input and external knowledge in a bottom-up manner, LLMs hold extensive, ready-to-use knowledge from pre-training. Consequently, recent efforts aim to direct LLMs to concentrate solely on pertinent knowledge for specific IE tasks.


\section{Table of Traditional OpenIE Models}
\label{appendix:milestone}

\begin{table*}[t]
\centering
\scalebox{0.55}{
\begin{tabular}{c|l|l|c|r|l}
\hline
Model & \multicolumn{1}{c|}{Method} & \multicolumn{1}{c|}{Source of Information} & \begin{tabular}[c]{@{}c@{}}Task\\ Setting\end{tabular} & \multicolumn{1}{c|}{Dataset} & \multicolumn{1}{c}{Evaluation (Result)} \\ \hline
\begin{tabular}[c]{@{}c@{}}\textsc{TextRunner}\\ \cite{banko2007open}\end{tabular} & \begin{tabular}[c]{@{}l@{}}Dependency Parser, NP Chunker,\\ CRF, Naive Bayes Classifier\end{tabular} & syntactic, dependency & ORTE & 400 Web & Average Error Rate (12\%) \\ \hline
\begin{tabular}[c]{@{}c@{}}WOE\\ \cite{wu2010open}\end{tabular} & \begin{tabular}[c]{@{}l@{}}\textsc{TextRunner},\\ Self-supervised Learning\end{tabular} & syntactic, dependency & ORTE & \begin{tabular}[c]{@{}r@{}}300 news\\ 300 Wikipedia\\ 300 Web\end{tabular} & Precision-Recall Curve \\ \hline
\begin{tabular}[c]{@{}c@{}}\textsc{ReVerb}\\ \cite{fader2011identifying}\end{tabular} & \begin{tabular}[c]{@{}l@{}}Syntactic Constraints,\\ Lexical Contraints, CRF\end{tabular} & syntactic, dependency & ORTE & 500 Web & \begin{tabular}[c]{@{}l@{}}Precision-Recall Curve,\\ AUC (1.3*$\textrm{WOE}^{parse}$, 2*\textsc{TextRunner})\end{tabular} \\ \hline
\begin{tabular}[c]{@{}c@{}}\textsc{Ollie}\\ \cite{schmitz2012open}\end{tabular} & \begin{tabular}[c]{@{}l@{}}\textsc{ReVerb}, Bootstrap, \\ Open Pattern Learning\end{tabular} & syntactic, dependency & ORTE & \begin{tabular}[c]{@{}r@{}}300 news (from WOE)\\ 300 Wikipedia (from WOE)\\ 300 biology\end{tabular} & \begin{tabular}[c]{@{}l@{}}Precision-Yield Curve,\\ AUC (1.9*$\textrm{WOE}^{parse}$, 2.7*\textsc{ReVerb})\end{tabular} \\ \hline
\begin{tabular}[c]{@{}c@{}}\textsc{OpenIE4}\\ \cite{mausam2016open}\end{tabular} & \begin{tabular}[c]{@{}l@{}}\textsc{SrlIE} \cite{christensen2011analysis}, \\ \textsc{RelNoun} \cite{pal2016demonyms}\end{tabular} & syntactic, dependency & ORTE & Not Reported & \begin{tabular}[c]{@{}l@{}}Precision-Yield Curve,\\ AUC (1.32*\textsc{Ollie}, 4*\textsc{ReVerb})\end{tabular} \\ \hline
\begin{tabular}[c]{@{}c@{}}ClausIE\\ \cite{del2013clausie}\end{tabular} & \begin{tabular}[c]{@{}l@{}}Dependency Parser,\\ Clause-based Model\end{tabular} & syntactic, dependency & ORTE & \begin{tabular}[c]{@{}r@{}}500 Web (from REVERB)\\ 200 Wikipedia\\ 200 news\end{tabular} & \begin{tabular}[c]{@{}l@{}}Precision-Yield Curve,\\ \# of correct extractions / \# of extractions\end{tabular} \\ \hline
\multirow{4}{*}{\begin{tabular}[c]{@{}c@{}}RnnOIE\\ \cite{Stanovsky2018SupervisedOI}\end{tabular}} & \multirow{4}{*}{Bi-LSTM, Softmax} & \multirow{4}{*}{word emb, POS emb} & \multirow{4}{*}{ORTE} & OIE2016 & AUC (48), F1 (62) \\
 &  &  &  & WEB & AUC (47), F1 (67) \\
 &  &  &  & NYT & AUC (25), F1 (35) \\
 &  &  &  & PENN & AUC (26), F1 (44) \\ \hline
\begin{tabular}[c]{@{}c@{}}NeuralOIE\\ \cite{Cui2018NeuralOI}\end{tabular} & LSTM, Copy Attention & word emb & ORTE & OIE2016 & AUC (27) \\ \hline
\begin{tabular}[c]{@{}c@{}}IMoJIE\\ \cite{Kolluru2020IMoJIEIM}\end{tabular} & BERT, LSTM, CopyAttention & word emb & ORTE & CaRB & AUC (33.3), F1 (53.5) \\ \hline
\multirow{2}{*}{\begin{tabular}[c]{@{}c@{}}SpanOIE\\ \cite{Zhan2020SpanMF}\end{tabular}} & \multirow{2}{*}{\begin{tabular}[c]{@{}l@{}}Bi-LSTM,\\ Span-consistent Greedy Search\end{tabular}} & \multirow{2}{*}{\begin{tabular}[c]{@{}l@{}}word emb, POS emb,\\ dependency relation emb\end{tabular}} & \multirow{2}{*}{ORTE} & OIE2016 & AUC (48.9), F1 (68.65) \\
 &  &  &  & Re-OIE2016 & AUC (65.9), F1 (78.50) \\ \hline
\multirow{2}{*}{\begin{tabular}[c]{@{}c@{}}$\textrm{Multi}^{2}\textrm{OIE}$\\ \cite{Ro2020Multi2OIEMO}\end{tabular}} & \multirow{2}{*}{BERT, Multihead Attention} & \multirow{2}{*}{\begin{tabular}[c]{@{}l@{}}word emb, position emb,\\ avg vector of predicates\end{tabular}} & \multirow{2}{*}{ORTE} & Re-OIE2016 & AUC (74.6), F1 (83.9) \\
 &  &  &  & CaRB & AUC (32.6), F1 (52.3) \\ \hline
\begin{tabular}[c]{@{}c@{}}OpenIE6\\ \cite{Kolluru2020OpenIE6IG}\end{tabular} & \begin{tabular}[c]{@{}l@{}}Iterative Grid Labeling,\\ BERT, Self-attention\end{tabular} & \begin{tabular}[c]{@{}l@{}}word emb,\\ dependency feature\end{tabular} & ORTE & CaRB & AUC (33.7), F1 (52.7) \\ \hline
\multirow{3}{*}{\begin{tabular}[c]{@{}c@{}}HNN4ORT\\ \cite{jia2022hybrid}\end{tabular}} & \multirow{3}{*}{ON-LSTM, CNN, Attention} & \multirow{3}{*}{\begin{tabular}[c]{@{}l@{}}word emb, POS emb,\\ argument emb,\\ local/global features\end{tabular}} & \multirow{3}{*}{ORTE} & Wikipedia & F1 (79.8) \\
 &  &  &  & NYT & F1 (74.5) \\
 &  &  &  & Reverb & F1 (81.7) \\ \hline
\begin{tabular}[c]{@{}c@{}}UORE\\ \cite{elsahar2017unsupervised}\end{tabular} & \begin{tabular}[c]{@{}l@{}}Re-weight Word Emb,\\ TF-IDF, PCA, HAC\end{tabular} & \begin{tabular}[c]{@{}l@{}}word emb,\\ dependency\end{tabular} & ORC & NYT-FB & F1 (41.6) \\ \hline
\begin{tabular}[c]{@{}c@{}}RSN\\ \cite{wu2019open}\end{tabular} & \begin{tabular}[c]{@{}l@{}}Relational Siamese Network,\\ CNN, HAC, Louvain\end{tabular} & word emb & ORC & FewRel & ${B}^{3}$: P (48.9) R (77.5) F1 (59.9) \\ \hline
\multirow{7}{*}{\begin{tabular}[c]{@{}c@{}}SelfORE\\ \cite{hu2020selfore}\end{tabular}} & \multirow{6}{*}{\begin{tabular}[c]{@{}l@{}}Bootstrapping Self-supervision,\\ BERT, K-means,\\ Adaptive Clustering\end{tabular}} & \multirow{7}{*}{word emb} & \multirow{3}{*}{ORC} & NYT+FB & \begin{tabular}[c]{@{}l@{}}ARI (40.3),\\ ${B}^{3}$: P (49.1) R (47.3) F1 (51.1),\\ V: F1 (46.6) Hom (45.7) Comp (47.6)\end{tabular} \\ \cline{5-6} 
 &  &  &  & T-REx SPO & \begin{tabular}[c]{@{}l@{}}ARI (33.7),\\ ${B}^{3}$: P (41.0) R (39.4) F1 (42.8),\\ V: F1 (41.4) Hom (40.3) Comp (42.5)\end{tabular} \\ \cline{5-6} 
 &  &  &  & T-REx DS & \begin{tabular}[c]{@{}l@{}}ARI (20.1), \\ ${B}^{3}$: P (32.9) R (29.7) F1 (36.8),\\ V: F1 (32.4) Hom (30.1) Comp (35.1)\end{tabular} \\ \hline
\multirow{4}{*}{\begin{tabular}[c]{@{}c@{}}OHRE\\ \cite{Zhang2021OpenHR}\end{tabular}} & \multirow{2}{*}{\begin{tabular}[c]{@{}l@{}}CNN, Virtual Adversarial Training,\\ Reconstruction Loss,\\ Dynamic Hierarchical Triplet Loss,\\ Louvain\end{tabular}} & \multirow{4}{*}{\begin{tabular}[c]{@{}l@{}}word emb,\\ hierarchical information\end{tabular}} & \multirow{4}{*}{ORC} & FewRel Hierarchy & \begin{tabular}[c]{@{}l@{}}ARI (64.2), \\ ${B}^{3}$: P (64.5) R (77.7) F1 (70.5),\\ V: F1 (76.7) Hom (73.8) Comp (79.9)\end{tabular} \\ \cline{5-6} 
 &  &  &  & NYT-FB Hierarchy & \begin{tabular}[c]{@{}l@{}}ARI (31.9), \\ ${B}^{3}$: P (31.4) R (72.3) F1 (43.8),\\ V: F1 (60.0) Hom (49.9) Comp (75.3)\end{tabular} \\ \hline
\multirow{4}{*}{\begin{tabular}[c]{@{}c@{}}ElementORE\\ \cite{Fangchao2021ElementIF}\end{tabular}} & \multirow{4}{*}{\begin{tabular}[c]{@{}l@{}}BERT, T5 \cite{Raffel2020ExploringTL},\\ Structure Causal Model, K-means\end{tabular}} & \multirow{4}{*}{\begin{tabular}[c]{@{}l@{}}word emb,\\ hierarchical information\end{tabular}} & \multirow{4}{*}{ORC} & T-REx SPO & \begin{tabular}[c]{@{}l@{}}ARI (36.6 ), \\ ${B}^{3}$: P (46.7) R (43.4) F1 (45.0),\\ V: F1 (45.3) Hom (45.4) Comp (45.2)\end{tabular} \\ \cline{5-6} 
 &  &  &  & T-REx DS & \begin{tabular}[c]{@{}l@{}}ARI (25.0), \\ ${B}^{3}$: P (40.2) R (45.9) F1 (42.9),\\ V: F1 (47.3) Hom (46.9) Comp (47.8)\end{tabular} \\ \hline
\begin{tabular}[c]{@{}c@{}}RoCORE\\ \cite{Zhao2021ARC}\end{tabular} & \begin{tabular}[c]{@{}l@{}}Relation-oriented Representation,\\ BERT, K-means\end{tabular} & word emb & ORC & FewRel & \begin{tabular}[c]{@{}l@{}}ARI (70.9), \\ ${B}^{3}$: P (75.2) R (84.6) F1 (79.6),\\ V: F1 (86.0) Hom (83.8) Comp (88.3)\end{tabular} \\ \hline
\multirow{4}{*}{\begin{tabular}[c]{@{}c@{}}\textsc{DeepEx}\\ \cite{wang2021zero}\end{tabular}} & \multirow{4}{*}{\begin{tabular}[c]{@{}l@{}}BERT, Attention, Beam Search,\\ Contrastive Pre-training\end{tabular}} & \multirow{4}{*}{\begin{tabular}[c]{@{}l@{}}NP chunks,\\ word emb, triplet emb\end{tabular}} & \multirow{4}{*}{ORSE} & OIE2016 & AUC (58.6), F1 (72.6) \\
 &  &  &  & WEB & AUC (82.4), F1 (91.2) \\
 &  &  &  & NYT & AUC (72.5), F1 (85.5) \\
 &  &  &  & PENN & AUC (81.5), F1 (88.5) \\ \hline
\end{tabular}
}
\caption{Milestone and representative models of pre-LLM era. ("V" denotes "V-measure", and "emb" stands for "embedding".)}
\label{tab:milestone}
\end{table*}

\end{document}